\documentclass[
]{ceurart}

\usepackage{listings}
\usepackage{hyperref}
\usepackage{url}
\begin{document}

\copyrightyear{2025}
\copyrightclause{Copyright for this paper by its authors.
  Use permitted under Creative Commons License Attribution 4.0
  International (CC BY 4.0).}

\conference{MiGA@IJCAI25: International IJCAI Workshop on 3rd Human Behavior Analysis for Emotion Understanding, August 29, 2025, Guangzhou, China.}

\title{Towards Fine-Grained Emotion Understanding via Skeleton-Based Micro-Gesture Recognition}


\author[1]{Hao Xu}[%
orcid=0009-0003-8156-1765,
email=2865346358@qq.com,
]

\author[1]{Lechao Cheng}[%
orcid=0000-0002-7546-9052,
email=chenglc@hfut.edu.cn,
]
\cormark[1]
\author[1]{Yaxiong Wang}[%
orcid=0000-0001-6596-8117,
email=wangyx15@stu.xjtu.edu.cn,
]

\author[1]{Shengeng Tang}[%
orcid=0000-0001-6313-2543,
email=tangsg@hfut.edu.cn,
]

\author[1]{Zhun Zhong}[%
orcid=0000-0002-8202-0544,
email=zhunzhong007@gmail.com,
]
\address[1]{Hefei University of Technology,
   No. 485, Danxia Road, Shushan District, Hefei, 230601, China}

\cortext[1]{Corresponding author.}

\begin{abstract}
We present our solution to the MiGA Challenge at IJCAI 2025, which aims to recognize micro-gestures (MGs) from skeleton sequences for the purpose of hidden emotion understanding. MGs are characterized by their subtlety, short duration, and low motion amplitude, making them particularly challenging to model and classify. We adopt PoseC3D as the baseline framework and introduce three key enhancements: (1) a topology-aware skeleton representation specifically designed for the iMiGUE dataset to better capture fine-grained motion patterns; (2) an improved temporal processing strategy that facilitates smoother and more temporally consistent motion modeling; and (3) the incorporation of semantic label embeddings as auxiliary supervision to improve the model’s generalization ability. Our method achieves a Top-1 accuracy of 67.01\% on the iMiGUE test set. As a result of these contributions, our approach ranks third on the official MiGA Challenge leaderboard. The source code is available at \href{https://github.com/EGO-False-Sleep/Miga25_track1}{https://github.com/EGO-False-Sleep/Miga25\_track1}.
\end{abstract}

\begin{keywords}
  Micro-gesture \sep
  action classification \sep
  data preprocessing \sep
  skeleton-based action recognition
\end{keywords}

\maketitle

\section{Introduction}

The Micro-Gesture Analysis for Hidden Emotion Understanding (MiGA) Challenge at IJCAI 2025 focuses on advancing the recognition of micro-gestures (MGs)—brief, involuntary body movements that convey subtle emotional cues. Unlike conventional action~\cite{lu2024mixed, lu2025mixed} or gesture recognition~\cite{tang2025sign, tang2025discrete}, MG recognition presents unique challenges due to the subtlety~\cite{zhang2025towards}, short duration, and low amplitude of the gestures, as well as the strong contextual dependency and significant class imbalance~\cite{fang2023separating} in the datasets. This challenge builds upon two publicly available datasets, iMiGUE~\cite{liu2021imigue} and SMG~\cite{chen2023smg}, with iMiGUE featuring spontaneous gestures collected from real-world post-match press conferences. To address these challenges, we formulate the MG classification task as a skeleton-based action recognition problem and adopt PoseC3D~\cite{duan2022revisiting} as our baseline model. We further introduce three key improvements: (1) a joint skeleton topology with a customized connectivity structure inspired by ST-GCN~\cite{yan2018spatial}, tailored to better capture localized motion patterns in MGs; (2) a refined temporal processing strategy that ensures smoother and more coherent skeletal motion representation; and (3) the integration of semantic label embeddings~\cite{frome2013devise, yeh2019multilabel, wei2020learning, filntisis2020emotion} to provide auxiliary supervision for improved generalization across imbalanced and semantically overlapping categories. Our final model achieves a Top-1 accuracy of 67.01\% on the iMiGUE test set, demonstrating the effectiveness of our topology-aware, semantically guided modeling framework for micro-gesture recognition.
The main contributions of our method are summarized as follows.
\begin{itemize}
\item We propose a joint graph structure inspired by ST-GCN~\cite{yan2018spatial}, along with a novel joint connectivity topology that more accurately captures the fine-grained motion patterns specific to the iMiGUE dataset. We also perform a cross-dataset analysis to investigate how different skeletal configurations influence recognition performance across architectures.

\item We introduce an improved temporal sampling and alignment strategy that departs from the original ST-GCN formulation. This approach enhances motion continuity and enables more coherent representation of raw skeleton sequences.

\item Our complete pipeline, including the proposed topological and temporal enhancements, achieves a Top-1 accuracy of 67.01\% on the iMiGUE test set.

\end{itemize}

\begin{figure}[!htbp]
    \centering
    \vspace{-3mm}
    \includegraphics[width=0.9\linewidth]{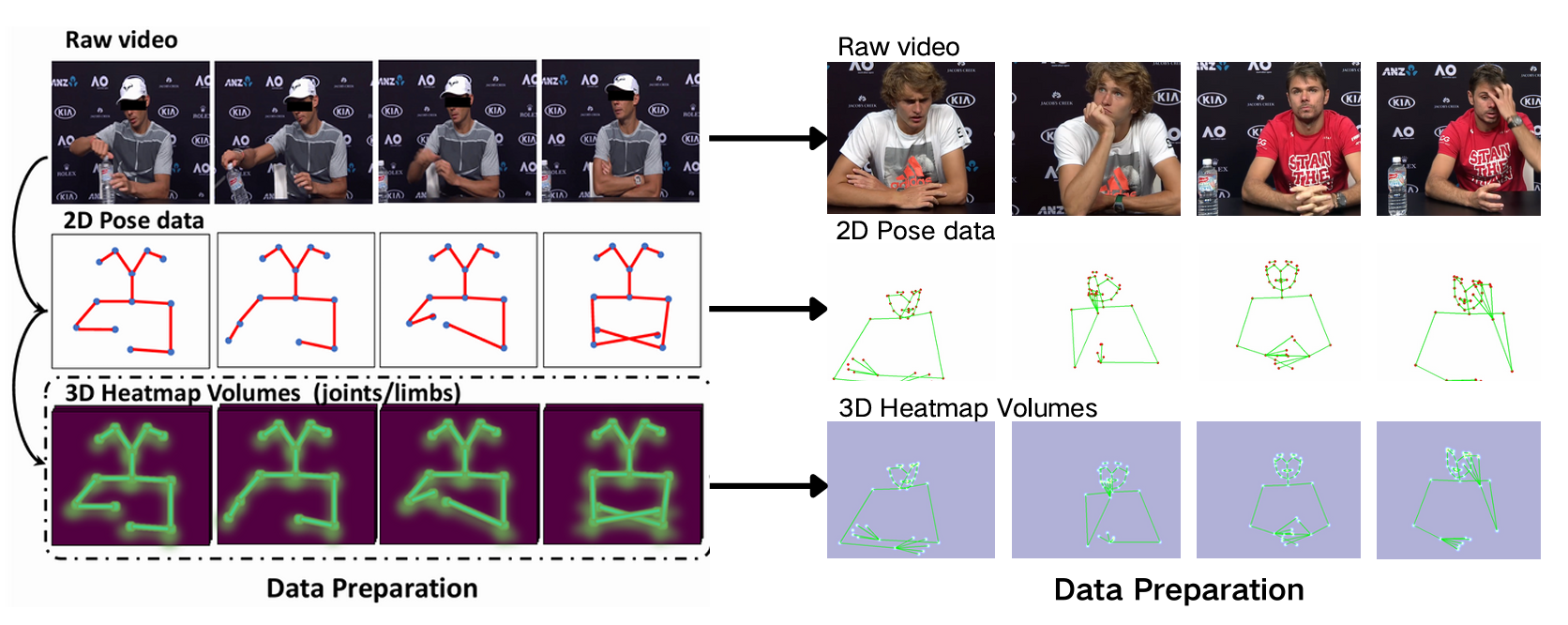}
    \caption{Visualization of the proposed new skeletal joint connection diagram for micro-gesture classification.}
    \label{fig:enter-label}
\end{figure}

\section{Methodology}

In this challenge, we experiment with two representative skeleton-based action recognition frameworks: {\bf ST-GCN} and {\bf PoseC3D}. This section details our design choices, empirical findings, and analysis based on both architectures.

\subsection{Skeleton Augmentation with Facial Keypoints}

Human skeletal connectivity is spatially consistent and relatively easy for graph-based models to learn. However, in the iMiGUE micro-gesture recognition task, many action categories—such as touching the face, adjusting a hat, or biting lips—are localized in the facial region. To enhance facial motion perception, we extend the standard 22-joint OpenPose skeleton to a 41-joint structure by incorporating additional facial landmarks (e.g., cheeks, eyebrows, and lips). This augmentation provides finer spatial resolution in regions critical to emotion-related micro-gestures.

Figure~\ref{fig:enter-label} compares the skeletal connectivity diagrams under different keypoint configurations. While this modification benefits representation, it also diverges from the original graph topology assumptions of ST-GCN, as discussed in the next subsection.

\begin{figure}[!htbp]
    \centering
    \includegraphics[width=0.8\linewidth]{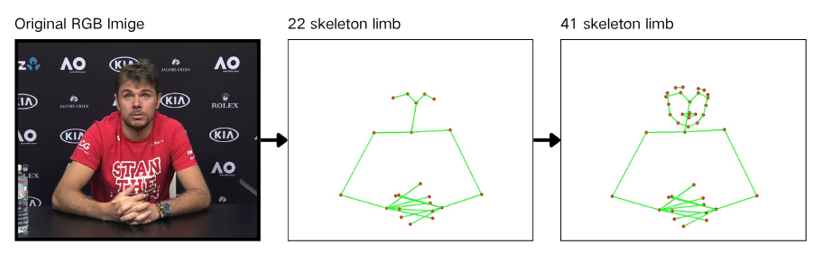}
    \caption{Comparison of skeletal connectivity with different numbers of keypoints.}
    \label{fig:enter-label}
\end{figure}

\subsection{Partitioning and Graph Reasoning in ST-GCN}

Effective action recognition requires modeling meaningful spatiotemporal motion patterns. ST-GCN achieves this by partitioning the skeleton graph into sub-regions, such as centripetal, centrifugal, and stationary limbs~\cite{yan2018spatial}. While effective for coarse-scale action categories, we find this partitioning suboptimal for micro-motion classification, likely due to the lack of distinguishable limb dynamics and limited data scale.

To compensate, we first enrich the model input with additional facial keypoints. These regions, while less dynamic in raw motion, are semantically aligned with many micro-gesture classes. However, empirical results reveal that these added keypoints degrade performance in ST-GCN. We hypothesize this is due to:
\begin{itemize}
    \item Overfitting from increased graph complexity and limited training data.
    \item The added nodes not contributing salient temporal or relational motion patterns that ST-GCN is designed to exploit.
    \item The peripheral nature and low motion magnitude of facial keypoints reducing their relative attention in the learned graph features.
\end{itemize}
Thus, despite their semantic relevance, these keypoints are possibly treated as noise within the ST-GCN's fixed topology and partitioning scheme\footnote{We acknowledge that this interpretation may be influenced by our limited experience with ST-GCN and time constraints during the challenge. We welcome future improvements from the community in this direction.}.

\subsection{PoseC3D with Extended Keypoints}

Given the limitations of ST-GCN, we shift our focus to PoseC3D, a 3D-CNN based method that processes skeletons as spatiotemporal heatmaps. Even in its baseline form, PoseC3D significantly outperforms ST-GCN. More importantly, when extended with facial keypoints, PoseC3D benefits substantially in performance. We attribute this to two key properties:

\begin{itemize}
    \item \textbf{Heatmap-based Representation}: PoseC3D encodes keypoints as dense heatmaps, which preserve richer spatial information and allow the network to infer latent movement patterns—even in low-motion regions. This representation has higher information entropy than raw joint coordinates, enabling stronger generalization.
    \item \textbf{Flexible 3D Convolutions}: The spatiotemporal convolutions in PoseC3D operate over the entire motion volume with uniform treatment of all locations. Unlike GCN's, the receptive field and feature propagation are not constrained by predefined skeletal graphs, granting PoseC3D greater expressivity and robustness to irrelevant noise.
\end{itemize}

\subsection{Temporal Frame Stream Processing}
Temporal modeling is a critical component in micro-gesture recognition, given the subtlety and brevity of such actions. The default temporal sampling strategy in ST-GCN adopts simple rule-based heuristics:

\begin{itemize}
\item When the number of frames exceeds the target length, a continuous subsequence is randomly cropped from the original sequence;
\item When the number of frames is insufficient, zero-padding is applied to extend the sequence to the required length.
\end{itemize}

However, these strategies often fail to preserve the complete temporal structure of micro-gestures. Random cropping may exclude key motion cues, while zero-padding introduces artificial discontinuities that disrupt temporal coherence. These limitations are particularly detrimental in micro-motion scenarios, where discriminative features are both sparse and temporally localized. To address these issues, we propose a structure-preserving temporal alignment strategy as follows:

\begin{itemize}
\item For over-length sequences, we perform uniform interval sampling, ensuring that both the first and last frames are retained. This guarantees that the sampled sequence spans the full temporal range of the original gesture;
\item For under-length sequences, we apply linear interpolation to generate intermediate frames, thereby expanding the sequence to the target length while maintaining temporal smoothness and continuity.
\end{itemize}

Compared to conventional approaches, the proposed strategy offers better coverage of the gesture trajectory and preserves fine-grained motion dynamics. Empirical results further confirm that this refinement contributes to improved model stability and recognition accuracy in the micro-gesture classification task.

\section{Experiments}

\subsection{Dataset: iMiGUE~\cite{liu2021imigue}}

The Micro-Gesture Understanding and Emotion Analysis (iMiGUE) dataset consists of 32 micro-gesture (MG) categories and one additional non-MG class. All data are collected from post-match press conference videos of professional tennis players. The dataset comprises a total of 18,499 annotated MG samples, which are labeled from 359 long video sequences ranging in duration from 0.5 to 26 minutes, totaling approximately 3,765,600 frames. iMiGUE provides two modalities for each sample: (1) RGB videos, and (2) 2D skeletal joint coordinates extracted using the OpenPose pose estimation framework. This multi-modal design enables both appearance-based and skeleton-based gesture analysis, supporting the development of robust models for emotion understanding based on subtle behavioral cues.

\subsection{Evaluation Metrics and Implementation Details}
We evaluate micro-gesture classification performance using the Top-1 Accuracy, which measures the percentage of samples for which the predicted label exactly matches the ground truth. Our method is implemented based on the open-source \texttt{PySkl} toolbox~\cite{duan2022pyskl}, and the training pipeline incorporates a loss function inspired by the winning solution of the \textit{MiGA 2023} challenge. The model is trained using Stochastic Gradient Descent (SGD) with a momentum of 0.9, a weight decay of $3 \times 10^{-4}$, and a batch size of 24. The initial learning rate is set to $0.1/3$, and we adopt a Cosine Annealing learning rate schedule. We use ResNet3D-SlowOnly as the feature extraction backbone and I3D as the classification head. For multi-stream ensemble modeling, which integrates joint and limb modalities, we apply a weighted fusion scheme with a ratio of 1:1, ensuring equal contribution from both sources of motion information.

 

\subsection{Experiments}

\subsection{Quantitative Results}

Table~\ref{tab:leaderboard} presents the top-3 results on the iMiGUE test set as reported on the official \texttt{CodaLab} competition leaderboard\footnote{Available at: \url{https://www.kaggle.com/competitions/the-3rd-mi-ga-ijcai-challenge-track-1/leaderboard}}. Our method achieves a Top-1 Accuracy of 67.01\%, ranking third overall among all participants.

\begin{table}[h]
\centering
\caption{Top-3 entries on the iMiGUE test set (Top-1 Accuracy).}
\label{tab:leaderboard}
\begin{tabular}{c|c|c}
\toprule
\textbf{Rank} & \textbf{Team} & \textbf{Top-1 Accuracy (\%)} \\
\midrule
1 & \texttt{gkdx2} & 73.213 \\
2 & \texttt{awuniverse} & 68.697 \\
3 & \textbf{Ours} & \textbf{67.010} \\
\bottomrule
\end{tabular}
\end{table}

\vspace{0.5em}

Table~\ref{tab:comparison} compares our method against standard PoseC3D and the winning solution of MiGA2023~\cite{li2023joint} across different input modalities on the iMiGUE dataset. All models use the same backbone (\texttt{ResNet3D-SlowOnly}) and classification head (\texttt{I3D}). Our method consistently outperforms existing approaches, achieving improvements across all modality models compared to the 2023 champion solution that also uses semantic embeddings.~4

\begin{table}[h]
\centering
\caption{Performance comparison under different modality settings on iMiGUE.}
\label{tab:comparison}
\begin{tabular}{l|c|c}
\toprule
\textbf{Method} & \textbf{Modality} & \textbf{Top-1 Accuracy (\%)} \\
\midrule
ST-GCN~\cite{yan2018spatial} & Joint & 46.38 \\
PoseC3D~\cite{duan2022revisiting} & Joint & 59.54 \\
PoseC3D~\cite{duan2022revisiting} & Limb & 60.74 \\
MiGA2023 1st~\cite{li2023joint} & Joint & 62.28 \\
MiGA2023 1st~\cite{li2023joint} & Limb & 63.48 \\
MiGA2023 1st~\cite{li2023joint} & Joint \& Limb & 64.12 \\
\midrule
\textbf{Ours} & Joint & \textbf{65.43 (+3.15)} \\
\textbf{Ours} & Limb & \textbf{64.40 (+0.92)} \\
\textbf{Ours} & Joint \& Limb & \textbf{67.01 (+2.89)} \\
\bottomrule
\end{tabular}
\end{table}

These results demonstrate the effectiveness of our approach in modeling subtle motion cues through enhanced skeleton topology, refined temporal processing, and semantic label supervision.





\section{Conclusions}
In this paper, we present the solution developed for the MiGA Challenge held at IJCAI 2025. Throughout the process, we employed both the ST-GCN and PoseC3D models, comparing their similarities and differences to explore the relationship between convolutional approaches and sequential data. Ultimately, by leveraging joint and limb modality data and adopting PoseC3D as the backbone—combined with the semantic embedding loss\cite{li2023joint} proposed in 2023—our method achieved third place with a Top-1 accuracy of 67.01\%. For this task, we recognize that there remains ample room for further research. Moving forward, we plan to address the challenges from additional perspectives, such as improved denoising techniques, strategies for handling imbalanced data, and the integration of RGB video streams, among others.

\section{Acknowledgements}
This work has been supported by the National Natural Science Foundation of China (Grant No. 62472139), by the Anhui Provincial Natural Science Foundation, China (Grant No. 2408085QF191).

\bibliography{sample-ceur}

\begin{thebibliography}{16}
\expandafter\ifx\csname natexlab\endcsname\relax\def\natexlab#1{#1}\fi
\providecommand{\url}[1]{\texttt{#1}}
\providecommand{\href}[2]{#2}
\providecommand{\path}[1]{#1}
\providecommand{\DOIprefix}{doi:}
\providecommand{\ArXivprefix}{arXiv:}
\providecommand{\URLprefix}{URL: }
\providecommand{\Pubmedprefix}{pmid:}
\providecommand{\doi}[1]{\href{http://dx.doi.org/#1}{\path{#1}}}
\providecommand{\Pubmed}[1]{\href{pmid:#1}{\path{#1}}}
\providecommand{\bibinfo}[2]{#2}
\ifx\xfnm\relax \def\xfnm[#1]{\unskip,\space#1}\fi
\bibitem[{Lu et~al.(2024)Lu, Zhao, Cheng, Zheng, Fan, and Song}]{lu2024mixed}
\bibinfo{author}{X.~Lu}, \bibinfo{author}{S.~Zhao}, \bibinfo{author}{L.~Cheng}, \bibinfo{author}{Y.~Zheng}, \bibinfo{author}{X.~Fan}, \bibinfo{author}{M.~Song},
\newblock \bibinfo{title}{Mixed resolution network with hierarchical motion modeling for efficient action recognition},
\newblock \bibinfo{journal}{Knowledge-Based Systems} \bibinfo{volume}{294} (\bibinfo{year}{2024}) \bibinfo{pages}{111686}.
\bibitem[{Lu et~al.(2025)Lu, Hao, Cheng, Zhao, Liu, and Song}]{lu2025mixed}
\bibinfo{author}{X.~Lu}, \bibinfo{author}{Y.~Hao}, \bibinfo{author}{L.~Cheng}, \bibinfo{author}{S.~Zhao}, \bibinfo{author}{Y.~Liu}, \bibinfo{author}{M.~Song},
\newblock \bibinfo{title}{Mixed attention and channel shift transformer for efficient action recognition},
\newblock \bibinfo{journal}{ACM Transactions on Multimedia Computing, Communications and Applications} \bibinfo{volume}{21} (\bibinfo{year}{2025}) \bibinfo{pages}{1--20}.
\bibitem[{Tang et~al.(2025{\natexlab{a}})Tang, He, Guo, Wei, Li, and Hong}]{tang2025sign}
\bibinfo{author}{S.~Tang}, \bibinfo{author}{J.~He}, \bibinfo{author}{D.~Guo}, \bibinfo{author}{Y.~Wei}, \bibinfo{author}{F.~Li}, \bibinfo{author}{R.~Hong},
\newblock \bibinfo{title}{Sign-idd: Iconicity disentangled diffusion for sign language production},
\newblock in: \bibinfo{booktitle}{Proceedings of the AAAI Conference on Artificial Intelligence}, volume~\bibinfo{volume}{39}, \bibinfo{year}{2025}{\natexlab{a}}, pp. \bibinfo{pages}{7266--7274}.
\bibitem[{Tang et~al.(2025{\natexlab{b}})Tang, He, Cheng, Wu, Guo, and Hong}]{tang2025discrete}
\bibinfo{author}{S.~Tang}, \bibinfo{author}{J.~He}, \bibinfo{author}{L.~Cheng}, \bibinfo{author}{J.~Wu}, \bibinfo{author}{D.~Guo}, \bibinfo{author}{R.~Hong},
\newblock \bibinfo{title}{Discrete to continuous: Generating smooth transition poses from sign language observations},
\newblock in: \bibinfo{booktitle}{Proceedings of the Computer Vision and Pattern Recognition Conference}, \bibinfo{year}{2025}{\natexlab{b}}, pp. \bibinfo{pages}{3481--3491}.
\bibitem[{Zhang et~al.(2025)Zhang, Cheng, Wang, Zhong, and Wang}]{zhang2025towards}
\bibinfo{author}{Y.~Zhang}, \bibinfo{author}{L.~Cheng}, \bibinfo{author}{Y.~Wang}, \bibinfo{author}{Z.~Zhong}, \bibinfo{author}{M.~Wang},
\newblock \bibinfo{title}{Towards micro-action recognition with limited annotations: An asynchronous pseudo labeling and training approach},
\newblock \bibinfo{journal}{arXiv preprint arXiv:2504.07785}  (\bibinfo{year}{2025}).
\bibitem[{Fang et~al.(2023)Fang, Cheng, Mao, Zhang, Fang, Li, Qi, and Jiao}]{fang2023separating}
\bibinfo{author}{C.~Fang}, \bibinfo{author}{L.~Cheng}, \bibinfo{author}{Y.~Mao}, \bibinfo{author}{D.~Zhang}, \bibinfo{author}{Y.~Fang}, \bibinfo{author}{G.~Li}, \bibinfo{author}{H.~Qi}, \bibinfo{author}{L.~Jiao},
\newblock \bibinfo{title}{Separating noisy samples from tail classes for long-tailed image classification with label noise},
\newblock \bibinfo{journal}{IEEE Transactions on Neural Networks and Learning Systems}  (\bibinfo{year}{2023}).
\bibitem[{Liu et~al.(2021)Liu, Shi, Chen, Yu, Li, and Zhao}]{liu2021imigue}
\bibinfo{author}{X.~Liu}, \bibinfo{author}{H.~Shi}, \bibinfo{author}{H.~Chen}, \bibinfo{author}{Z.~Yu}, \bibinfo{author}{X.~Li}, \bibinfo{author}{G.~Zhao},
\newblock \bibinfo{title}{imigue: An identity-free video dataset for micro-gesture understanding and emotion analysis},
\newblock in: \bibinfo{booktitle}{Proceedings of the IEEE/CVF conference on computer vision and pattern recognition}, \bibinfo{year}{2021}, pp. \bibinfo{pages}{10631--10642}.
\bibitem[{Chen et~al.(2023)Chen, Shi, Liu, Li, and Zhao}]{chen2023smg}
\bibinfo{author}{H.~Chen}, \bibinfo{author}{H.~Shi}, \bibinfo{author}{X.~Liu}, \bibinfo{author}{X.~Li}, \bibinfo{author}{G.~Zhao},
\newblock \bibinfo{title}{Smg: A micro-gesture dataset towards spontaneous body gestures for emotional stress state analysis},
\newblock \bibinfo{journal}{International Journal of Computer Vision} \bibinfo{volume}{131} (\bibinfo{year}{2023}) \bibinfo{pages}{1346--1366}.
\bibitem[{Duan et~al.(2022)Duan, Zhao, Chen, Lin, and Dai}]{duan2022revisiting}
\bibinfo{author}{H.~Duan}, \bibinfo{author}{Y.~Zhao}, \bibinfo{author}{K.~Chen}, \bibinfo{author}{D.~Lin}, \bibinfo{author}{B.~Dai},
\newblock \bibinfo{title}{Revisiting skeleton-based action recognition},
\newblock in: \bibinfo{booktitle}{Proceedings of the IEEE/CVF conference on computer vision and pattern recognition}, \bibinfo{year}{2022}, pp. \bibinfo{pages}{2969--2978}.
\bibitem[{Yan et~al.(2018)Yan, Xiong, and Lin}]{yan2018spatial}
\bibinfo{author}{S.~Yan}, \bibinfo{author}{Y.~Xiong}, \bibinfo{author}{D.~Lin},
\newblock \bibinfo{title}{Spatial temporal graph convolutional networks for skeleton-based action recognition},
\newblock in: \bibinfo{booktitle}{Proceedings of the AAAI conference on artificial intelligence}, volume~\bibinfo{volume}{32}, \bibinfo{year}{2018}.
\bibitem[{Frome et~al.(2013)Frome, Corrado, Shlens, Bengio, Dean, Ranzato, and Mikolov}]{frome2013devise}
\bibinfo{author}{A.~Frome}, \bibinfo{author}{G.~S. Corrado}, \bibinfo{author}{J.~Shlens}, \bibinfo{author}{S.~Bengio}, \bibinfo{author}{J.~Dean}, \bibinfo{author}{M.~Ranzato}, \bibinfo{author}{T.~Mikolov},
\newblock \bibinfo{title}{Devise: A deep visual-semantic embedding model},
\newblock \bibinfo{journal}{Advances in neural information processing systems} \bibinfo{volume}{26} (\bibinfo{year}{2013}).
\bibitem[{Yeh and Li(2019)}]{yeh2019multilabel}
\bibinfo{author}{M.-C. Yeh}, \bibinfo{author}{Y.-N. Li},
\newblock \bibinfo{title}{Multilabel deep visual-semantic embedding},
\newblock \bibinfo{journal}{IEEE transactions on pattern analysis and machine intelligence} \bibinfo{volume}{42} (\bibinfo{year}{2019}) \bibinfo{pages}{1530--1536}.
\bibitem[{Wei et~al.(2020)Wei, Zhang, Lin, Lee, Balasubramanian, Hoai, and Samaras}]{wei2020learning}
\bibinfo{author}{Z.~Wei}, \bibinfo{author}{J.~Zhang}, \bibinfo{author}{Z.~Lin}, \bibinfo{author}{J.-Y. Lee}, \bibinfo{author}{N.~Balasubramanian}, \bibinfo{author}{M.~Hoai}, \bibinfo{author}{D.~Samaras},
\newblock \bibinfo{title}{Learning visual emotion representations from web data},
\newblock in: \bibinfo{booktitle}{Proceedings of the IEEE/CVF Conference on Computer Vision and Pattern Recognition}, \bibinfo{year}{2020}, pp. \bibinfo{pages}{13106--13115}.
\bibitem[{Filntisis et~al.(2020)Filntisis, Efthymiou, Potamianos, and Maragos}]{filntisis2020emotion}
\bibinfo{author}{P.~P. Filntisis}, \bibinfo{author}{N.~Efthymiou}, \bibinfo{author}{G.~Potamianos}, \bibinfo{author}{P.~Maragos},
\newblock \bibinfo{title}{Emotion understanding in videos through body, context, and visual-semantic embedding loss},
\newblock in: \bibinfo{booktitle}{Computer Vision--ECCV 2020 Workshops: Glasgow, UK, August 23--28, 2020, Proceedings, Part I 16}, \bibinfo{organization}{Springer}, \bibinfo{year}{2020}, pp. \bibinfo{pages}{747--755}.
\bibitem[{Duan et~al.(2022)Duan, Wang, Chen, and Lin}]{duan2022pyskl}
\bibinfo{author}{H.~Duan}, \bibinfo{author}{J.~Wang}, \bibinfo{author}{K.~Chen}, \bibinfo{author}{D.~Lin},
\newblock \bibinfo{title}{Pyskl: Towards good practices for skeleton action recognition},
\newblock in: \bibinfo{booktitle}{Proceedings of the 30th ACM International Conference on Multimedia}, \bibinfo{year}{2022}, pp. \bibinfo{pages}{7351--7354}.
\bibitem[{Li et~al.(2023)Li, Guo, Chen, Peng, and Wang}]{li2023joint}
\bibinfo{author}{K.~Li}, \bibinfo{author}{D.~Guo}, \bibinfo{author}{G.~Chen}, \bibinfo{author}{X.~Peng}, \bibinfo{author}{M.~Wang},
\newblock \bibinfo{title}{Joint skeletal and semantic embedding loss for micro-gesture classification},
\newblock \bibinfo{journal}{arXiv preprint arXiv:2307.10624}  (\bibinfo{year}{2023}).

\end{thebibliography}
\end{document}